\documentclass[letterpaper, 10 pt, conference]{ieeeconf}  

\IEEEoverridecommandlockouts                              

\usepackage{float}
\usepackage{amsmath} 
\usepackage{amssymb}  
\usepackage{multicol}
\usepackage{graphicx}
\usepackage{multirow}
\usepackage{booktabs}  
\usepackage{cuted}
\usepackage{capt-of}

\title{\LARGE \bf
PhyVisGen: Physically and Visually High-Fidelity Robotic Manipulation Data Generation
}

\author{Yu Zheng$^{1*}$, Qiyu Feng$^{1*}$, Yixin Wu$^{1*}$, Baoquan Yang$^{1}$, Yixuan Zhou$^{3}$, Bingyang Hu$^{2}$, \\Kemeng Huang$^{4,2}$, Guansheng Yang$^{2}$ and Hesheng Wang$^{1}$
\thanks{*The first three authors contributed equally. 
Corresponding Author: Hesheng Wang, Guansheng Yang (e-mail: wanghesheng@sjtu.edu.cn).
}
\thanks{$^{1}$School of Automation and Intelligent Sensing, Shanghai Jiao Tong University, Shanghai 200240, China.
}
\thanks{$^{2}$Shenzhen Dizhou Technology Co., Ltd., Shenzhen, China} 
\thanks{$^{3}$Global College, Shanghai Jiao Tong University, Shanghai, China.
}
\thanks{$^{4}$The University of Hong Kong, Hong Kong, China.} 
}

\begin{document}

\maketitle
\thispagestyle{empty}
\pagestyle{empty}

\begin{abstract}
Large-scale manipulation demonstrations are essential for learning robust visuomotor policies, yet real-world data collection is expensive and difficult to scale. Simulation offers a promising alternative, but physical and visual discrepancies can limit the transferability of synthetic data, particularly for manipulation with soft grippers. We present PhyVisGen, a physically and visually high-fidelity framework for scalable robotic manipulation data generation. On the physical side, PhyVisGen introduces an arm-gripper coupling method based on the Incremental Potential Contact (IPC), enabling high-fidelity soft contact throughout complete manipulation trajectories. On the visual side, it combines real-scene reconstruction with real-time path tracing to generate visually realistic observations while preserving captured scene appearance. Quantitative evaluations demonstrate the physical and visual fidelity of PhyVisGen. Policies trained exclusively on synthetic manipulation demonstrations achieve 65–95\% success across five real-robot tasks, without real-robot demonstration data or policy fine-tuning.

\end{abstract}

\section{INTRODUCTION}

Robot manipulation increasingly relies on imitation learning~\cite{act,dp}, which learns complex skills from expert demonstrations. However, achieving robust generalization typically requires large-scale data, while collecting demonstrations on physical robots is expensive and labor-intensive.

To extend the manipulation data scale, a growing body of work~\cite{robotwin,mimicplay,rialto,re3sim,jiang2025dexmimicgen,skillmimic,lodestar} has explored simulation-based approaches for automatically generating robot manipulation data. 
However, scaling the quantity of synthetic demonstrations alone does not ensure their fidelity to real-world manipulation. For contact-rich manipulation learning, discrepancies arise from both the physical interactions and the visual observations.

On the physical side, recent manipulation simulators have incorporated deformable-body models, substantially improving the simulation of deformable objects~\cite{huang2021plasticinelab,chen2023daxbench,saghour2024,huang2022defgraspsim,moghani2026softmimicgen}. However, deformability in real robotic systems is not limited to manipulated objects: soft grippers also deform during contact, continuously changing the contact geometry and force transmission~\cite{chi2024universal}. Incremental Potential Contact (IPC)-based methods~\cite{kim2022ipc,zema2024,ma2025grip} have enabled high-fidelity simulation of soft gripper-object interactions, but existing approaches mainly focus on isolated grasp generation, execution, or stability evaluation. Extending high-fidelity soft contact from isolated grasps to complete manipulation trajectories remains underexplored.

On the visual side, existing real-to-sim approaches~\cite{re3sim,robosimgs,ROLA} commonly rely on reconstructed scene representations or image compositing to preserve the appearance of real environments. While these approaches can reproduce realistic backgrounds, rasterization or visual blending cannot faithfully model geometry-dependent light transport between simulated objects and the reconstructed scene, including shadows, reflections, and refractions. Ray-tracing-based rendering engine~\cite{isaacsim,sapien} provides more physically consistent light interactions, but directly re-rendering reconstructed textured meshes introduces another challenge: their textures already encode the illumination observed during real-world capture, while the renderer applies additional illumination, leading to appearance shifts and lighting artifacts.

\begin{figure}[t]
	\centering
    \resizebox{1.\linewidth}{!}
	{
		\includegraphics[scale=1.0]{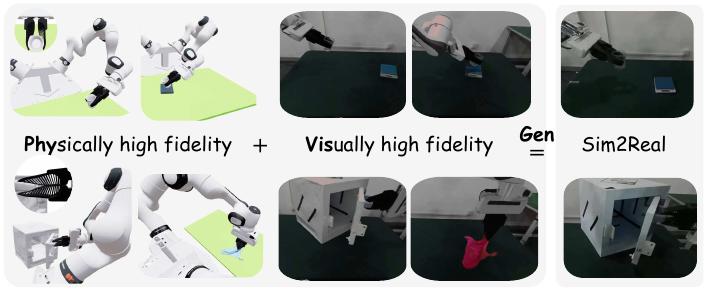}
        }
	\caption{\textbf{PhyVisGen} generates the manipulation data with the interaction between the soft gripper and rigid and deformable objects. Our IPC-based simulation ensures the physically high fidelity of the manipulation trajectory (\textbf{left}), and the path-tracing-based rendering realizes the visually high fidelity of the rendered image (\textbf{middle}). The generated data finally enables the policy to be zero-shot deployed in the real world (\textbf{right}).}
    
	\vspace{-3mm}
	\label{fig:head}
\end{figure}

To address the above challenges, we propose PhyVisGen, a physically and visually high-fidelity data generation framework for robot manipulation with soft grippers. The central goal of PhyVisGen is to reduce the sim-to-real gap from two complementary perspectives. 
On the physical side, we build on StiffGIPC~\cite{huang2025stiffgipc} and propose an arm-gripper coupling method that integrates robot arm links, rigid gripper components, and deformable fingertips within a unified incremental-potential formulation. 
This coupling allows contact forces on the soft fingertips to propagate back to the robot arm while jointly resolving robot motion, soft-gripper deformation, contact, friction, and object response. 
Thus, high-fidelity IPC-based soft-gripper simulation is extended from isolated grasping to complete manipulation trajectories.

On the visual side, we develop a real-time path tracer that efficiently simulates complex lighting and optical phenomena, such as shadows and transparency. More importantly, our appearance-preserving shadow-receiving material enables reconstructed meshes to be directly placed into the simulator while preserving their original real-world appearance, while still supporting physically consistent shadow interactions with simulated objects. We further introduce a channel-wise 1D lookup table (LUT) calibration to compensate for rendering and tone-mapping discrepancies.

We evaluate PhyVisGen on a range of soft-gripper manipulation tasks. Experimental results show that visuomotor policies trained solely on data synthesized by PhyVisGen can be deployed to real robotic systems without any fine-tuning on real robot data and successfully perform multiple zero-shot sim-to-real manipulation tasks.

In summary, the main contributions are as follows:

\begin{itemize}
    \item We propose PhyVisGen, a physically and visually high-fidelity data generation framework for robot manipulation with soft grippers. 
    \item We build a physically high-fidelity soft-gripper manipulation system, realizing arm-gripper coupling and complete trajectory rollout.
    \item We develop a visually high-fidelity renderer, which enables physically consistent lighting interactions with simulated objects while preserving the captured appearance of reconstructed real-world scenes.
    \item We demonstrate that manipulation policies trained solely on synthetic data generated by PhyVisGen can achieve zero-shot sim-to-real deployment across multiple soft-gripper manipulation tasks.
\end{itemize}

\section{RELATED WORK}

\subsection{Manipulation Data Generation}
Conventional data collection pipelines typically rely on teleoperation~\cite{act,chi2024universal}, which provides high-quality real-world data but is difficult to scale across large numbers of tasks, objects, and scene configurations. To improve data efficiency, several works have explored simulation-based demonstration generation~\cite{robotwin,mimicplay,rialto,re3sim}.
Among them, MimicGen~\cite{mimicplay} decomposes a few source demonstrations into object-centric trajectories and adapts them to novel object locations, enabling large-scale demonstration synthesis. 
Related efforts have further extended automated demonstration generation to bimanual manipulation, dexterous manipulation, and long-horizon task manipulation~\cite{jiang2025dexmimicgen,lodestar,skillmimic}.

However, most existing methods primarily emphasize the scale and diversity of generated trajectories, while relying on simplified contact models or conventional rendering pipelines. As a result, the physical and visual discrepancies between simulation and reality can remain substantial, particularly for manipulation involving soft grippers or deformable objects. Our work complements this line of research by focusing on both the fidelity of physical interactions and visual observations during scalable manipulation data generation.

\subsection{Physical Simulation}
Deformable-object simulation has been increasingly adopted in robotic manipulation to capture interaction dynamics that cannot be adequately represented by rigid-body models. Among existing approaches, Material Point Method (MPM)-based simulation has been widely explored for deformable-object manipulation~\cite{huang2021plasticinelab,chen2023daxbench}. In parallel, Finite Element Method (FEM)-based approaches provide high-fidelity modeling for deformable object manipulation~\cite{saghour2024,huang2022defgraspsim}.
These works substantially improve the modeling and data generation capability for manipulation involving deformable objects.

Beyond deformable objects, soft grippers have been increasingly adopted in robotic manipulation for their improved contact adaptability and tolerance to geometric and pose uncertainties. Their deformation, however, fundamentally alters contact geometry and force transmission, placing higher demands on contact modeling and simulation fidelity.
IPC-based methods~\cite{kim2022ipc,zema2024,ma2025grip} provide high-fidelity simulation of contact and deformation between soft grippers and manipulated objects. IPC-GraspSim~\cite{kim2022ipc} explicitly models deformable fingertip–object interactions and demonstrates improved grasp prediction accuracy. GRIP~\cite{ma2025grip} further accelerates IPC-based simulation to enable large-scale generation of deformable–rigid coupled grasping data. 

Despite these advances, existing soft-gripper simulators are largely designed for isolated grasp generation, execution, or stability evaluation. 
Less attention has been paid to integrating the soft gripper with manipulation tasks. 
Our work incorporates high-fidelity soft contact into robot manipulation processes and using the resulting high-fidelity trajectories for policy learning.

\begin{figure*}[t]
	\centering
    \resizebox{1.\linewidth}{!}
	{
		\includegraphics[scale=1.0]{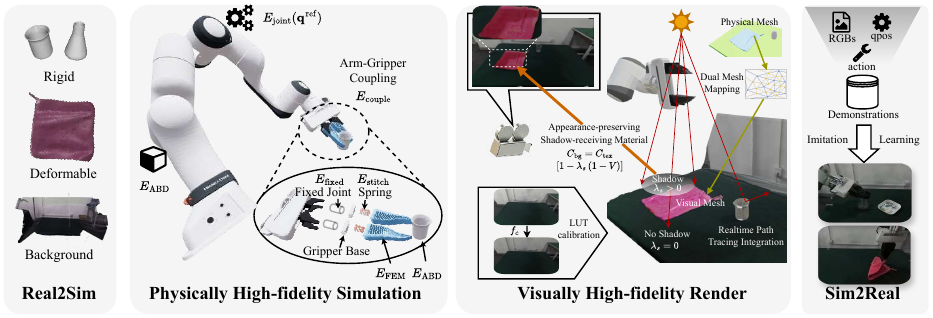}
        }
	\caption{\textbf{Pipeline of PhyVisGen}. The real2sim provides the textured reconstruction of the manipulated objects and background. The reconstructed objects are first fed into the IPC-based and arm-gripper-coupled simulation, which generates the physically high-fidelity trajectories. The trajectories are input into the rendering system, which adopts path tracing, appearance-preserved shadow-receiving material, dual mesh mapping, and LUT calibration to generate the visually high-fidelity observations. The demonstrations are used for imitation learning. The learned policies are zero-shot deployed in the real world.  }
	\vspace{-3mm}
	\label{fig:main}
\end{figure*}

\subsection{Realistic Rendering}
Performing 3D reconstruction of real-world scenes and rendering interactive high-fidelity observations is critical for narrowing the gap between simulation and reality in robot policy learning.
Existing methods fall into three categories. 
The first~\cite{robogs,splatsim,Rl-gsbridge} employs 3DGS for reconstruction and rendering. 
The second~\cite{ROLA} uses visual blending by compositing simulator-rendered results with background images.
While both of these approaches can faithfully preserve the background, methods relying solely on rasterization or visual blending cannot accurately model light visibility, shadow, reflection, and refraction between objects and the reconstructed background.
The third~\cite{isaacsim,sapien} leverages ray-tracing-based rendering engines.
However, directly rendering a reconstructed textured mesh introduces a fundamental appearance inconsistency: the texture already encodes the illumination observed during real-world capture, while the rendering engine applies additional illumination during re-rendering, resulting in undesired reflections and other lighting artifacts. Moreover, even when such illumination-induced discrepancies are suppressed, the rendered images can still exhibit photometric differences from real camera observations due to discrepancies in camera response and image formation~\cite{LUT1}.

In contrast, our work combines reconstructed real-world scenes with physically based rendering while explicitly addressing the appearance inconsistencies introduced when re-rendering captured textures, enabling more faithful visual interactions between simulated objects and reconstructed environments.

\section{METHOD}

In this section, we will first introduce the framework of our PhyVisGen, followed by the details of each component in the pipeline.
\subsection{Framework}

As illustrated in Fig.~\ref{fig:main}, PhyVisGen consists of real2sim reconstruction, physically and visually high-fidelity demonstration generation, and zero-shot sim2real deployment. We first reconstruct sim-ready assets from the real-world setup. Our coupled arm-gripper simulator then generates complete manipulation trajectories with physically high fidelity. The physical states are transferred to a path-tracing pipeline to produce visually realistic observations while preserving the reconstructed scene appearance. The demonstrations are used to train visuomotor policies that are directly deployed on the real robot without real-world fine-tuning.

\subsection{Real2Sim}
We reconstruct the objects and the background separately.
For rigid and deformable object reconstruction, we adopt PGSR~\cite{PGSR} for opaque objects and TSGS~\cite{TSGS} for transparent ones, and then further extract the reconstructed GS into watertight, simulation-ready mesh assets.
For scene background, we use Map-Anything~\cite{mapanything} to reconstruct a metric-scale scene mesh from RGB-D inputs.
\subsection{Physically High-Fidelity Simulation}
\subsubsection{Soft Gripper Simulation}
We consider a manipulation system consisting of the rigid robotic arm, the soft gripper, and manipulated objects that can be either rigid or deformable. 
We explicitly model the soft gripper as a deformable body and resolve gripper–object interactions using a high-fidelity contact solver.

To robustly simulate contact under large deformation, we adopt StiffGIPC~\cite{huang2025stiffgipc} as the underlying contact formulation. 
StiffGIPC represents contact using barrier potentials together with continuous collision detection, preventing interpenetration while maintaining stable contact under large deformation and sustained interaction.

We represent each soft fingertip as a volumetric deformable body discretized by a tetrahedral mesh. Its elastic response is modeled using the Stable Neo-Hookean constitutive model~\cite{smith2018stable}. Material parameters are specified by Young's modulus $E$ and Poisson's ratio $\nu$ for the soft fingertip.
Deformable objects are modeled within the same deformable-body formulation when required.
Rigid and near-rigid objects are 
modeled as Affine Body Dynamics (ABD)~\cite{abd} governed by a stiff orthogonality energy that keeps the affine transformation close to a rotation.
Therefore, rigid and fully deformable bodies coexist in the same variational system under a single contact, friction, and coupling model. 
This representation enables the simulator to handle both deformable-rigid and deformable-deformable interactions under a common contact model.

For soft fingertips, the deformation state directly determines the local contact interface. As the gripper closes, the fingertips deform around the object, increasing the effective contact area and redistributing contact forces.

\subsubsection{Arm-Gripper Coupling and Manipulation Rollout}
High-fidelity contact simulation alone is insufficient for generating complete robot manipulation trajectories. During manipulation, the soft gripper must continuously interact with the robot arms, controller, manipulated objects, and surrounding environment. 
We therefore couple the deformable gripper simulation with the robotic arm to form a unified manipulation system.

The arm and the soft gripper are mechanically connected through two levels of coupling. The finger base and the rigid gripper base are connected by a fixed joint.
The rigid gripper base is further attached to the FEM soft fingertip through a set of springs. Therefore, the coupling energy is:
\begin{equation}
E_{\mathrm{couple}}
=
E_{\mathrm{fixed}}
+
E_{\mathrm{stitch}},
\label{eq:couple_energy}
\end{equation}
where the stitch-spring energy is
\begin{equation}
E_{\mathrm{stitch}}
=
\frac{1}{2} k_c
\sum_{i\in\mathcal{A}}
\left\|
\mathbf{x}_{i,t}
-
\hat{\mathbf{x}}_{i,t}
\right\|_2^2,
\label{eq:stitch_energy}
\end{equation}
where $\mathcal{A}$ denotes the set of attachment vertices, $\mathbf{x}_{i,t}$ is the position of a FEM vertex on the soft fingertip, and $\hat{\mathbf{x}}_{i,t}$ is the corresponding attachment point carried by the simulated rigid gripper base. Since both sides of the springs participate in the same solution step, the coupling is bidirectional rather than a prescribed positional constraint.

Based on the arm-gripper coupling, we formulate the complete manipulation simulation system. 
At each simulation step, all degrees of freedom are jointly advanced by minimizing a single incremental potential:
\begin{equation}
\begin{aligned}
E_{\mathrm{total}} =\;&
E_{\mathrm{inertia}} + E_{\mathrm{barrier}}
+ E_{\mathrm{friction}} + \\
+& E_{\mathrm{FEM}}
+ E_{\mathrm{ABD}}
+ E_{\mathrm{joint}}(\mathbf{q}^{\mathrm{ref}}) 
+ E_{\mathrm{couple}}
 .
\end{aligned}
\label{eq:total_energy}
\end{equation}
Here, $E_{\mathrm{inertia}}$ is the inertial term arising from implicit time integration. $E_{\mathrm{barrier}}$ is the log-barrier contact potential, and $E_{\mathrm{friction}}$ is the smoothly mollified friction potential, together yielding intersection-free contact following the IPC formulation. $E_{\mathrm{FEM}}$ models the elastic deformation of soft components, and $E_{\mathrm{ABD}}$ enforces the rigid-body behavior of the ABD-represented robot links. $E_{\mathrm{joint}}$ models the actuated revolute and prismatic joints, where $\mathbf{q}^{\mathrm{ref}}$ is the
joint actuation target. $E_{\mathrm{couple}}$ is the coupling energy.

All degrees of freedom are advanced
jointly by minimizing this single incremental potential at each time
step. The reference joint trajectory $\mathbf{q}^{\mathrm{ref}}$ enters
the system only through $E_{\mathrm{joint}}$ as quadratic drive
potentials for the revolute and prismatic joints, analogous to position drives in a PD controller. Importantly,
$\mathbf{q}^{\mathrm{ref}}$ is not imposed kinematically. Instead, the realized robot-link poses are jointly solved together with the deformable states and contact interactions. The resulting coupling is therefore fully bidirectional, allowing contact forces acting on the soft fingertips to propagate through the gripper structure back to the robot arm.

This unified formulation is rolled out throughout the complete manipulation trajectory. Given the joint-reference sequence generated by the motion planner, the solver jointly resolves robot dynamics, soft-gripper deformation, contact evolution, friction, and object response at every timestep, thereby maintaining physically consistent coupling throughout the entire manipulation process.

\subsubsection{Motion Generation}

Based on the physically high-fidelity simulator described above, we adopt a rule-based planning strategy to generate complete manipulation trajectories. 
Each manipulation task is represented as a sequence of key motion elements.
The target poses and local manipulation trajectories in each motion element are transformed based on the object pose for generating the demonstration data under different object initial poses. The poses and trajectories are converted into robot commands through inverse kinematics and the robot motion planner. 

For each successful rollout, the simulator records the complete physical state. 
The resulting physical trajectories are:
\begin{equation}
\tau_{\mathrm{phy}}
=
\{
\mathbf{q}_t,
\mathbf{a}_t,
\mathbf{T}^{\mathrm{link}}_t,
\mathbf{T}^{\mathrm{rigid}}_t,
\mathbf{X}^{\mathrm{deform}}_t,
\mathbf{X}^{\mathrm{gripper}}_t
\}_{t=1}^{T},
\end{equation}
where $t$ is the timestep, $\mathbf{q}_t$ is the robot joint configuration, $\mathbf{a}_t$ executed action. $\mathbf{T}^{\mathrm{link}}_t$ are the poses of the robot links, $\mathbf{T}^{\mathrm{rigid}}_t$ represents the poses of rigid objects. $\mathbf{X}^{\mathrm{deform}}_t$ and $\mathbf{X}^{\mathrm{gripper}}_t$ denote the vertex positions of deformable objects and soft grippers, respectively.

\subsection{Visual High-fidelity Rendering}

Based on physical trajectory $\tau_{\mathrm{phy}}$ generated through physical simulator, the frame-wise RGB observations are rendered. Specifically, for each timestep $t$, the recorded rigid-body poses and deformable vertex states are transferred to their corresponding visual representations, and the RGB observation $\mathbf{o}_t$ is generated as
\begin{equation}
\mathbf{o}_t
=
\mathcal{R}
\left(
\mathbf{T}^{\mathrm{link}}_t,
\mathbf{T}^{\mathrm{rigid}}_t,
\mathbf{X}^{\mathrm{deform}}_t,
\mathbf{X}^{\mathrm{gripper}}_t;
\mathcal{C},
\mathcal{A}
\right),
\end{equation}
where $\mathcal{R}(\cdot)$ denotes the rendering pipeline, $\mathcal{C}$ represents the calibrated camera parameters, and $\mathcal{A}$ denotes the reconstructed visual assets obtained from the real2sim process.

To improve visual fidelity and rendering efficiency, we design the rendering pipeline $\mathcal{R}(\cdot)$ with four components: physics-to-render dual-mesh mapping for efficient simulation, a real-time path tracer for shadows and transparency, an appearance-preserving shadow-receiving material for preserving reconstructed background appearance, and channel-wise photometric calibration for reducing rendering-to-real discrepancies.

\subsubsection{Physics-to-Render Dual Mesh Mapping}
Dense meshes are computationally expensive for physical simulation, while overly simplified meshes degrade rendering quality. Therefore, we adopt a dual-mesh representation consisting of a coarse mesh $M_c$ for physical simulation and a fine mesh $M_f$ for visual rendering. The two meshes share the same motion states, while independently satisfying the requirements of physics and rendering.

For rigid objects, we use PAMO~\cite{pamo} to simplify the fine mesh $M_f$ into a coarse mesh $M_c$ for physical simulation, and directly transfer the simulated pose of $M_c$ to $M_f$ during rendering. For deformable objects, where vertex-wise deformations cannot be represented by a rigid pose, we obtain $M_f$ by applying $L$-level Loop subdivision to $M_c$. Since Loop subdivision is linear, the complete subdivision process can be precomputed as a sparse mapping
\begin{equation}
\mathbf{W}_{c\rightarrow f}
=
\mathbf{S}^{(L)}\cdots\mathbf{S}^{(1)}
\in\mathbb{R}^{N_f\times N_c},
\end{equation}
where $N_c$ and $N_f$ denote the numbers of vertices in $M_c$ and $M_f$, respectively. At runtime, the visual vertex states are obtained by
\begin{equation}
\mathbf{X}_t^f
=
\mathbf{W}_{c\rightarrow f}\mathbf{X}_t^c,
\end{equation}
where $\mathbf{X}_t^c$ and $\mathbf{X}_t^f$ are the coarse and fine vertex states at timestep $t$. 

\subsubsection{Realtime Path Tracing Integration}
We design a real-time rendering engine that achieves interactive frame rates while supporting full path tracing with multiple area lights and transparent materials. 
The physics engine provides rigid-body poses and soft-body vertex states at each timestep, while the rendering engine takes these dynamic states as input for path tracing. Our implementation achieves 24 FPS at $640\times360$ resolution with 32 SPP on a single NVIDIA RTX 5090.

\subsubsection{Appearance-preserving Shadow-receiving Material}

Let $M_{bg}$ denote the background mesh and $C_{\mathrm{tex}}$ its texture. Since $C_{\mathrm{tex}}$ already encodes illumination and shadows from real observations, directly treating it as albedo causes  a second illumination. Conversely, an emissive material preserves the appearance but cannot receive shadows from newly introduced objects.

To address this problem, we design an appearance-preserving shadow-receiving material that decouples background appearance from shadow interaction. The background color is defined as
\begin{equation}
{C}_{\mathrm{bg}}
=
C_{\mathrm{tex}}
\left[
1-\lambda_s(1-V)
\right],
\label{eq:bg_color}
\end{equation}
where $V\in[0,1]$ denotes the light visibility at the background surface point, estimated by ray tracing, and $\lambda_s\in[0,1]$ controls the shadow strength. Specifically, $\lambda_s=0$ fully preserves the reconstructed appearance, while $\lambda_s=1$ applies the full estimated shadow attenuation. In practice, $V$ is estimated by tracing shadow rays toward $N$ light samples drawn from the emitters' importance distribution and averaging their visibility.

The material does not evaluate a BRDF or accumulate direct, ambient, or indirect illumination. Instead, paths terminate upon hitting the background, ensuring that the reconstructed appearance is preserved while allowing objects to cast shadows onto the reconstructed geometry.

\subsubsection{LUT calibration}

Although the appearance-preserving material preserves the reconstructed appearance, tone mapping of the path tracer's HDR output can introduce residual color and tone discrepancies with real observations. To address these photometric differences, we introduce a channel-wise 1D lookup table (LUT) calibration method.

Given the reconstructed background texture $C_{\mathrm{tex}}$, we render the background mesh under the same camera viewpoint and scene configuration as the corresponding real observation, obtaining $\mathbf{I}_{\mathrm{render}}$. We then establish dense pixel-wise correspondences between $\mathbf{I}_{\mathrm{render}}$ and the corresponding real camera frame $\mathbf{I}_{\mathrm{GT}}$. After converting both images from sRGB to linear RGB, the corresponding pixel intensities are used to learn a channel-wise photometric mapping:
\begin{equation}
\begin{aligned}
I_{\mathrm{GT}}^{(c)}(p)
&=
f_c\left(
I_{\mathrm{render}}^{(c)}(p)
\right),
I_{\mathrm{render}}^{(c)}(p)
&=
\mathcal{R}(C_{\mathrm{tex}}),
\end{aligned}
\label{eq:lut_mapping}
\end{equation}
where $p$ denotes an image pixel, $c\in\{R,G,B\}$ denotes the color channel, 
$\mathcal{R}(\cdot)$ denotes the rendering operator, and 
$f_c:[0,1]\rightarrow[0,1]$ is a channel-wise 256-point 1D LUT.

To provide sufficient resolution for modeling the nonlinear photometric response in linear RGB space, we use 256 intensity bins, allowing the mapping to smoothly capture both low-light and highlight regions during appearance transfer. To suppress local fluctuations in the estimated response, we further enforce a non-decreasing constraint using the weighted Pool Adjacent Violators Algorithm (PAVA). The resulting channel-wise response functions are combined into a $256\times3$ LUT. During subsequent rendering, the calibrated LUT is applied independently to each color channel of the linear RGB output of every rendered frame, producing the final photometrically calibrated observations.

\begin{figure*}[t]
	\centering
    \resizebox{1.\linewidth}{!}
	{
		\includegraphics[scale=1.0]{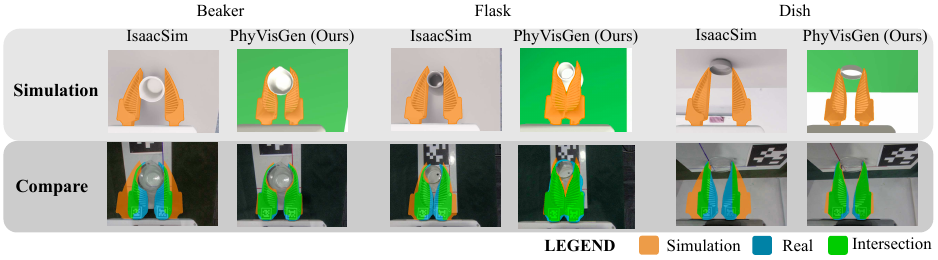}
        }
	\caption{The qualitative result of the physical fidelity evaluation. The \textbf{top} row demonstrates the observations and gripper masks from the simulation wrist camera in IsaacSim and our PhyVisGen. The \textbf{bottom} row shows the observations and gripper masks from the real-world wrist camera, as well as the comparison between the simulation and real-world gripper mask. }
	\label{fig:exp1}
\end{figure*}

\subsection{Sim2Real}
The physically and visually high-fidelity demonstrations $\mathcal{D}_{\mathrm{sim}}
    = \{\epsilon_i\}_{i=1}^{N}$ are used to train vision-based imitation learning policies
$\pi_\theta$, where each demonstration consists of the observation-action pairs, i.e., $\epsilon_i=\{(o_t,a_t)\}_t^T$. In general, the
training objective can be written as
\begin{equation}
    \theta^{*}
    =
    \arg\min_{\theta}
    \mathbb{E}_{(o_t,a_t)\sim\mathcal{D}_{\mathrm{sim}}}
    \left[
        \mathcal{L}_{\mathrm{IL}}
        \bigl(\pi_{\theta}(o_t), a_t\bigr)
    \right],
\end{equation}
where $\mathcal{L}_{\mathrm{IL}}$ denotes the policy-specific imitation
learning objective.
After training, the learned policy is directly deployed on the real
robot, resulting in a zero-shot
sim-to-real manipulation pipeline.

\section{EXPERIMENTS}
\label{sec:experiments}

In this section, we conduct extensive qualitative and quantitative evaluations to answer three key questions:
\begin{itemize}
    \item \textbf{Q1 (Physics Fidelity):} Does our physically high-fidelity simulator accurately reproduce the deformation of the real soft gripper during manipulation tasks?
    \item \textbf{Q2 (Visual Fidelity):} Does our visually high-fidelity renderer deliver rendering while accurately reproducing complex physical lighting and preserving background color fidelity?
    \item \textbf{Q3 (Sim-to-Real Transfer):} Do the generated data enable zero-shot real-world manipulation policy transfer?
\end{itemize}

\subsection{Robot Platform}
Our real-world experiments are conducted on a dual-arm robotic platform consisting of two Franka Research 3 manipulators and a ZED camera for visual observation. Each arm is equipped with a soft gripper based on the UMI~\cite{chi2024universal} gripper design. The grippers are fabricated from TPU using 3D printing and mounted on the respective robot arms.

\subsection{Physical Fidelity}\label{sec:pfs}
\subsubsection{Experiment Setting}

We establish a shared coordinate system between the simulation and the real robot using AprilTag~\cite{olson2011tags} markers to evaluate physical fidelity under aligned initial conditions and robot motions. 
We execute the same motion sequence generated by the motion-generation pipeline in both environments.

To quantitatively evaluate the physical fidelity of compliant gripper deformation during manipulation, we introduce the gripper mask IoU metric. 
We use a wrist camera to capture the soft gripper throughout the real-world manipulation process and place a virtual camera with the same calibrated pose in simulation to render the corresponding observations. For each pair of frames, we extract the gripper masks from the real and simulated images and compute their intersection-over-union (IoU) as
\begin{equation}
\mathrm{IoU}_t =
\frac{
\left|\mathcal{M}^{\mathrm{real}}_t \cap
\mathcal{M}^{\mathrm{sim}}_t\right|
}{
\left|\mathcal{M}^{\mathrm{real}}_t \cup
\mathcal{M}^{\mathrm{sim}}_t\right|
},
\end{equation}
where $\mathcal{M}^{\mathrm{real}}_t$ and $\mathcal{M}^{\mathrm{sim}}_t$ denote the gripper masks in the real and simulated images at timestep $t$, respectively. We report the mean IoU (mIoU) over the manipulation trajectory with the soft gripper deformed as the physical-fidelity measure, which measures how closely the simulated gripper deformation matches observations in the real world throughout the manipulation process.

We compare our physically high-fidelity simulator against Isaac Sim~\cite{isaacsim}. 
While Isaac Sim provides general-purpose deformable-body simulation, it does not provide an off-the-shelf model for a deformable soft gripper integrated with a robotic articulation. 
Accordingly, in our Isaac Sim baseline, the gripper is modeled as rigid and remains undeformed during interaction.
We conduct the fidelity evaluation on three fragile real-world objects, including the beaker, flask and dish, that require soft gripper during manipulation. 

\begin{table}[t]
\centering
\caption{The physical fidelity Comparison}
\resizebox{0.7\linewidth}{!}{
\begin{tabular}{lcc}
\toprule
\multirow{2}{*}{\textbf{Objects}} & \multicolumn{2}{c}{\textbf{mIoU (\%) }} \\
& Isaac Sim & PhyVisGen (Ours)\\
\midrule
Beaker & 39.49 & \textbf{83.34} \\ 
Flask & 45.16 & \textbf{82.97}    \\ 
Dish & 55.68 & \textbf{84.70}  \\

\bottomrule 
\end{tabular}
}
\label{tab:psc}
\vspace{-3mm}
\end{table}

\begin{figure*}[t]
	\centering
    \resizebox{0.9\linewidth}{!}
	{
		\includegraphics[scale=1.0]{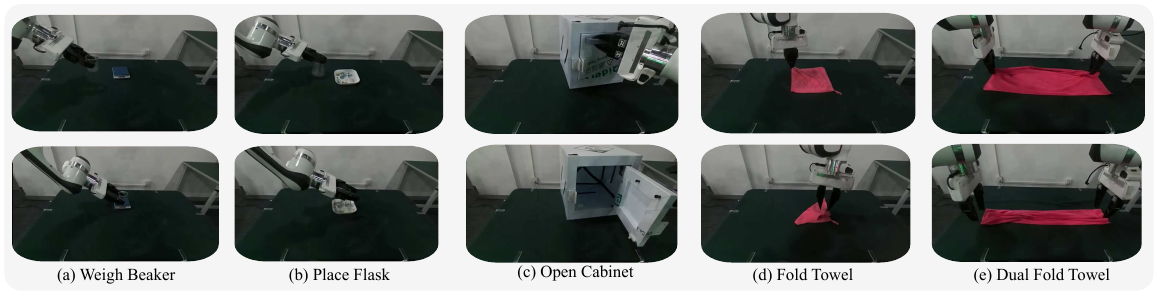}
        }
	\caption{\textbf{Sim-to-Real Policy Evaluation.} The example real-world trials of the five soft-gripper manipulation tasks with the ACT policies trained by the demonstrations generated by PhyVisGen. }
	\vspace{-3mm}
	\label{fig:exp3}
\end{figure*}

\begin{figure}[t]
	\centering
    \resizebox{0.9\linewidth}{!}
	{
		\includegraphics[scale=1.0]{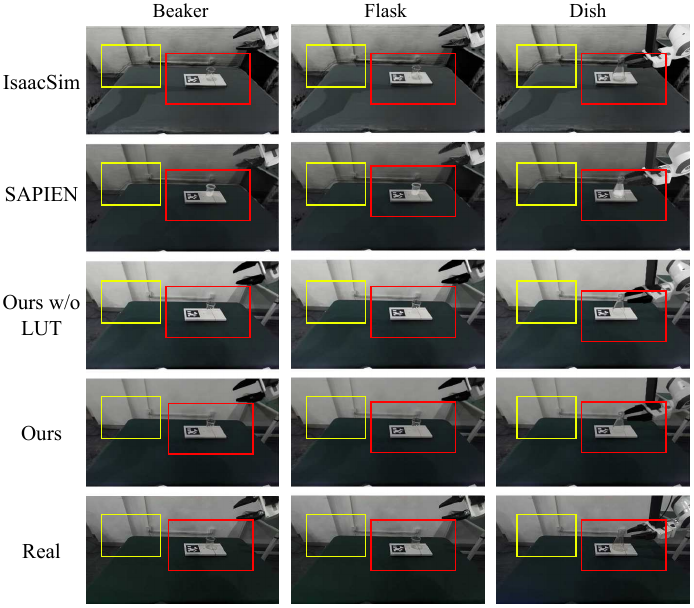}
        }
	\caption{\textbf{Visualization results of rendered and real images.} The details are highlighted in red and yellow boxes.}
	\label{fig:exp2}
\end{figure}

\subsubsection{Results}

As shown in Table~\ref{tab:psc}, the soft-gripper deformation produced by our simulator achieves mIoUs over 80\% across all three objects, indicating a close match to the real-world observations. In contrast, the mIoUs of Isaac Sim are below 60\%, exhibiting a substantially larger discrepancy from the real system. These results demonstrate that our IPC-based soft-gripper simulator can accurately reproduce the deformation behavior of the real soft gripper, thereby enabling the generation of physically high-fidelity manipulation trajectories.

The qualitative comparison in Fig.~\ref{fig:exp1} further supports this observation. The overlap region between our simulated gripper and the real gripper covers most of the gripper geometry, with only minor non-overlapping areas. In comparison, the Isaac Sim exhibits substantially larger mismatched regions. This qualitative result further demonstrates that our simulator faithfully reproduces real-world soft-gripper deformation.

\subsection{Visual Fidelity}
\subsubsection{Experiment Setting}
We use the same sim-real aligned data collected in Sec.~\ref{sec:pfs}. The RGB observations of the real ZED camera and RGB images rendered with the simulated camera are used for visual fidelity comparison.
For the quantitative comparison, we use PSNR, LPIPS, and SSIM over the images of the three objects as image quality metrics. Both the quantitative and qualitative comparisons are conducted against commonly used renderers, including SAPIEN~\cite{sapien} and Isaac Sim~\cite{isaacsim}.

\subsubsection{Results}

As shown in Table~\ref{tab:renderer_comparison}, our rendering method achieves the highest overall similarity to real-world observations. Even without LUT calibration, PhyVisGen achieves substantially better SSIM and LPIPS than both SAPIEN and Isaac Sim, indicating that our rendering pipeline can more faithfully reproduce the appearance of the real scene. Using the LUT calibration further improves the rendering consistency, achieving the best performance.

Beyond these quantitative metrics, Fig.~\ref{fig:exp2} provides qualitative comparisons. We show the comparison of the three objects respectively. As shown in the red box, our method renders transparent objects with more accurate light and shadow effects than existing approaches. The yellow box demonstrates that the appearance-preserving material eliminates secondary reflections caused by external illumination. Moreover, comparing Ours with Ours w/o LUT shows that the LUT calibration brings the rendered color tone closer to the real observation.

\begin{table}[t]
\centering
\caption{Rendering Performance Comparison. \textbf{Bold} and \underline{underlined} values indicate the best and second-best performance.}
\setlength{\tabcolsep}{10pt}
\resizebox{0.9\linewidth}{!}{
\begin{tabular}{lccc}
\toprule
& \textbf{PSNR} $\uparrow$ & \textbf{SSIM} $\uparrow$ & \textbf{LPIPS} $\downarrow$ \\
\midrule
SAPIEN      & \underline{18.45} & 0.80 & 0.25 \\
Isaac Sim   & 16.40 & 0.62 & 0.40 \\
Ours w/o LUT & 17.31 & \underline{0.82} & \underline{0.22} \\
Ours   & \textbf{23.87} & \textbf{0.88} & \textbf{0.17} \\
\bottomrule
\end{tabular}
}
\vspace{-3mm}
\label{tab:renderer_comparison}
\end{table}

\subsection{Sim-to-Real Policy Evaluation}

\subsubsection{Experiment Setting}

As shown in Fig.~\ref{fig:exp3}, we evaluate the sim-to-real capability of the data generated by PhyVisGen on five soft-gripper manipulation tasks beyond isolated grasping, including typical tasks needing the soft gripper and covering different object properties:
\begin{itemize}
    \item \textbf{Weigh Beaker}: The robot grasps the transparent beaker with the soft gripper and puts it on the scale.
    \item  \textbf{Place Flask}: The robot grasps the transparent flask with the soft gripper and puts it on the plate.
    \item \textbf{Open Cabinet}: The robot uses the soft gripper to tightly grasp the cabinet handle and follows a curve to open it.
    \item \textbf{Fold Towel}: The robot grasps one corner of the towel and folds it to the other corner.
    \item \textbf{Dual Fold Towel}: The robot uses the dual arms to grasp both sides of the big towel and fold it. 
\end{itemize}

For each task, we collect synthetic demonstrations using the PhyVisGen pipeline. We use the synthetic dataset to train the imitation learning policy Action Chunking with Transformers (ACT)~\cite{act}. 
After training, each policy is deployed directly on the real robot without any additional adaptation. We evaluate policy performance using task success rate over 20 real-world trials for each task.

\begin{table}[t]
\centering
\caption{The zero-shot real-world success rate of policies}
\resizebox{0.8\linewidth}{!}{
\setlength{\tabcolsep}{15pt}
\begin{tabular}{lc}
\toprule
Task & \textbf{Success Rate (\%)}\\
\midrule
\textit{Weigh Beaker}
& 90  \\ 
\textit{Place Flask} & 95    \\ 
\textit{Open Cabinet} & 80  \\
\textit{Fold Towel} & 65  \\
\textit{Dual Fold Towel} & 90  \\

\bottomrule 
\end{tabular}
}
\label{tab:real}
\vspace{-4mm}
\end{table}

\subsubsection{Results}
Table~\ref{tab:real} reports the zero-shot real-world transfer performance of ACT policies across five manipulation tasks. For the two tasks involving fragile transparent objects, our physical simulator captures the deformation behavior of the soft gripper around the objects, while the rendering pipeline reproduces the optical properties of transparent objects together with their shadow interactions with the reconstructed background, resulting in success rates at least 90\%. For the \textit{Open Cabinet} task, our simulator models the compliant deformation of the soft gripper around the cabinet handle, and the learned policy achieves an 80\% real-world success rate. Our framework can also manipulate deformable objects, achieving success rates of 65\% and 90\% on the two corresponding tasks. Overall, these results demonstrate that PhyVisGen can generate physically and visually high-fidelity manipulation demonstrations that are sufficient for training imitation-learning policies entirely in simulation and directly deploying them on real robots without using real-world training data.

\section{CONCLUSION}
We presented PhyVisGen, a physically and visually high-fidelity data generation framework for robotic manipulation with soft grippers. PhyVisGen introduces an arm-gripper coupling method within an IPC-based simulation framework, extending high-fidelity soft-contact simulation from isolated grasps to complete manipulation trajectories. It further integrates reconstructed real-world assets with real-time path tracing and an appearance-preserving shadow-receiving material to generate visually realistic observations while preserving the captured scene appearance. By combining these physical and visual components, PhyVisGen enables scalable generation of complete manipulation demonstrations for visuomotor policy learning. Our experiments show that policies trained solely on the generated synthetic data can be directly deployed on real robots without real-world fine-tuning. 

These results demonstrate the feasibility of using high-fidelity synthetic data for zero-shot sim-to-real soft-gripper manipulation.




\bibliographystyle{IEEEtran}
\bibliography{IEEEabrv,reference}

\end{document}